\pdfoutput=1
\documentclass[11pt]{article}

\usepackage[preprint]{acl}

\usepackage{times}
\usepackage{latexsym}

\usepackage[T1]{fontenc}
\usepackage[utf8]{inputenc}

\usepackage{microtype}

\usepackage{inconsolata}

\usepackage{graphicx}

\usepackage{booktabs}
\usepackage{tabularx}
\usepackage{makecell}
\usepackage{pifont}
\usepackage{arydshln}

\usepackage{tcolorbox}
\usepackage{multicol}
\usepackage{fvextra}

\title{\mlee: Evaluating Agents on Setting Up and Executing Tasks\\ from Research Repositories}

\author{\makecell{
  Ben Bogin$^{1,2}$~~~Kejuan Yang$^{2}$~~~Shashank Gupta$^{1}$~~~Kyle Richardson$^{1}$~~~Erin Bransom$^{1}$\\
  Peter Clark$^{1}$~~~~~Ashish Sabharwal$^{1}$~~~~~Tushar Khot$^{1}$} \\
  \hspace{1ex} \\
  $^{1}$Allen Institute for AI\hspace{5mm}
  $^{2}$University of Washington \\
  \texttt{\small\makecell{\{benb, shashankg, kyler, erinbransom, peterc, ashishs, tushark\}@allenai.org}}}

\usepackage[capitalize]{cleveref}
\crefformat{section}{\S#2#1#3}
\crefformat{subsection}{\S#2#1#3}
\crefformat{subsubsection}{\S#2#1#3}

\definecolor{DarkGreen}{RGB}{80,134,84}
\newcommand{\cmark}{\textcolor{DarkGreen}{\ding{51}}}
\newcommand{\xmark}{\textcolor{red}{\ding{55}}}

\usepackage{xspace}
\newcommand{\mlee}{\texttt{SUPER}\xspace}

\newcommand{\mlereact}{ReAct-\mlee}
\newcommand{\mleefull}{\textbf{S}etting \textbf{UP} and \textbf{E}xecuting tasks from \textbf{R}esearch repositories}

\newcommand{\expert}{Expert\xspace}
\newcommand{\scenario}{Masked\xspace}
\newcommand{\autogen}{Auto\xspace}
\newcommand{\numtasks}{45\xspace}
\newcommand{\numscenarios}{152\xspace}
\newcommand{\numautogen}{604\xspace}

\newif\ifcomments
\commentstrue 
\ifcomments
    \newcommand\tbd[1]{\textcolor{red}{TODO [#1]}}
    \newcommand\bb[1]{\textcolor{red}{[BB: #1]}}
    \newcommand\sg[1]{\textcolor{blue}{[SG: #1]}}
    \newcommand\tushar[1]{\textcolor{purple}{[TK: #1]}}
    \newcommand\ashish[1]{\textcolor{brown}{[AS: #1]}}
    \newcommand\kyle[1]{\textcolor{teal}{[KR: #1]}}
\else
    \providecommand[\tbd][1]{}
    \providecommand{\bb}[1]{}
    \providecommand{\sg}[1]{}
    \providecommand[\tushar][1]{}
    \providecommand{\ashish}[1]{}
    \providecommand{\kyle}[1]{}
\fi

\RecustomVerbatimCommand{\VerbatimInput}{VerbatimInput}{fontsize=\footnotesize,
 breaklines=true,
 frame=single,  
 framesep=0.5em, % separation between frame and text
 labelposition=topline,
}

\begin{document}
\maketitle

\begin{abstract}
Given that Large Language Models (LLMs) have made significant progress in writing code, 
can they now be used to autonomously reproduce results from research repositories?
Such a capability would be a boon to the research community, helping researchers validate, understand, and extend prior work. 
To advance towards this goal, we introduce \mlee, the first benchmark designed to evaluate the capability of LLMs in setting up and executing tasks from research repositories. \mlee aims to capture the realistic challenges faced by researchers working with Machine Learning (ML) and Natural Language Processing (NLP) research repositories. Our benchmark comprises three distinct problem sets: \numtasks \textit{end-to-end problems} with annotated expert solutions, \numscenarios \textit{sub-problems} derived from the expert set that focus on specific challenges (e.g., configuring a trainer), and \numautogen \textit{automatically generated problems} for larger-scale development. We introduce various evaluation measures to assess both task success and progress, utilizing gold solutions when available or approximations otherwise.
We show that state-of-the-art approaches struggle to solve these problems with the best model (GPT-4o) solving only 16.3\% of the end-to-end set, and 46.1\% of the scenarios. This illustrates the challenge of this task, and suggests that \mlee can serve as a valuable resource for the community to make and measure progress.\footnote{\url{https://github.com/allenai/super-benchmark}}
\end{abstract}
\section{Introduction}
Research and scientific discoveries often rely on the reproducibility of experiments conducted by other researchers and the ease with which scientists can build upon each other's work. In the context of empirical ML and NLP research, it is often crucial for researchers to be able to execute and reproduce diverse research experiments from open-source repositories, whether to verify existing results or to test them under new conditions. 

\begin{figure}[!t]
    \centering
  \includegraphics[width=0.97\linewidth]{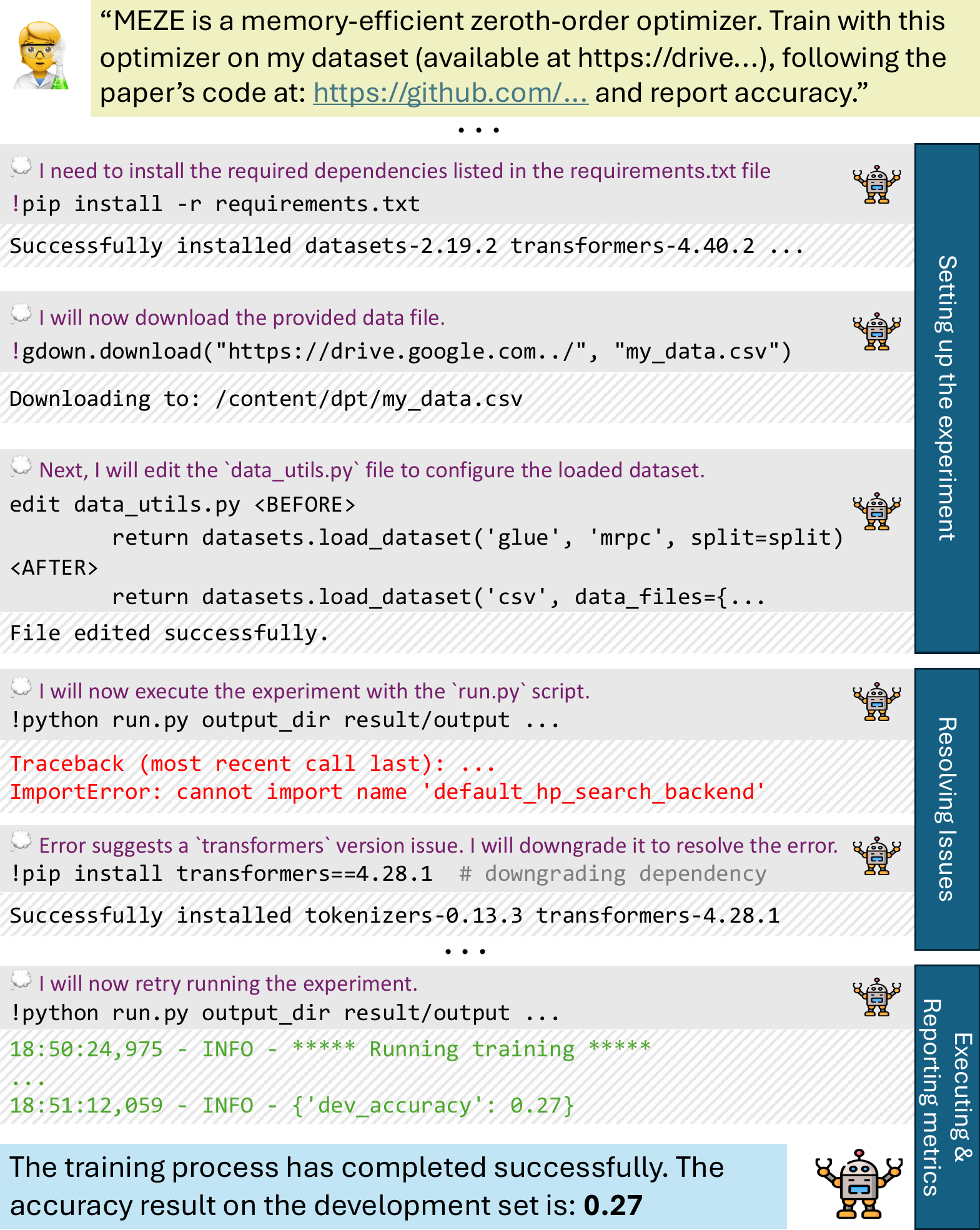}
    \caption{\small
    An illustration of a research task and some of the steps an agent would need to complete it, including updating data loading configuration, resolving dependency issues (due to unlisted version dependencies in the repository), running the training script and reporting metrics.
    }
    \label{fig:example}
\end{figure}

In practice, even when research code is available, running code from arbitrary repositories is often non-trivial and time-consuming \cite{Samuel2022ComputationalRO,storks-etal-2023-nlp}. Experimentation frequently requires substantial effort to \textit{set up and execute} them: installing the environment, making non-trivial configuration changes, resolving outdated package dependencies, fixing bugs, and determining the correct execution commands, among other tasks. All of this requires a considerable understanding of the documentation and repository code, knowledge about fixing issues (e.g., CUDA errors), as well as the ability to modify the code appropriately. These steps are especially time-consuming for research repositories ``in-the-wild'', as support and documentation may not be available.

In this work, we ask: \emph{Can LLMs automate the set up and execution of tasks in research repositories?} Consider the research task in \cref{fig:example} where the agent is asked to use a research code repository
to train a model with a new optimizer, and evaluate its performance on a custom dataset. A successful agent would need to set up the experiment by installing dependencies, downloading the provided data, and making code changes to load it (first three cells in the figure), then execute the training script while responding to unexpected issues such as an incompatible dependency (fourth and fifth cell), and finally report the result metrics (last cell).

While LLM-based agents have recently been used to produce execution commands from popular research repositories \cite{mlbench}, execute popular ML repositories \cite{mlagentbench}, or resolve repository issues \cite{swebench}, no existing benchmark evaluates agents on the common problem faced by many researchers: both \textit{setting up} and \textit{executing} experiments using research repositories \textit{in-the-wild}, i.e., less popular repositories that are not typically well-documented or maintained, which make experiments harder to configure and execute. As a recent study shows \cite{storks-etal-2023-nlp}, both novice and advanced researchers find the challenge of ``setting up the code base'' to be the most difficult part of reproducing experiments.

To encourage research on this problem, we introduce \textbf{\mlee} (\mleefull), a benchmark focusing on such lower-profile research repositories. \mlee consists of three distinct problem sets. The \textit{\expert} set contains \numtasks manually curated problems solved by experts. The \textit{\scenario} set includes \numscenarios sub-problems derived from the expert set through our proposed ``Code Masking'' mechanism, where we remove parts of the expert-written code to generate a diverse set of sub-problems targeting specific challenges. Each sub-problem addresses a specific challenge, such as installing dependencies and resolving conflicts, configuring experimental data, setting up hyper-parameters, resolving runtime exceptions, correctly executing scripts, etc. Lastly, the \emph{\autogen} set contains an additional \numautogen automatically generated tasks with an even more diverse set of repositories and challenges. It can potentially be used in future work for development, fine-tuning, or training using environment feedback.

To evaluate agents on the \expert and \scenario sets, for which we have gold solutions, we compare their answers (e.g., metrics to be reported) to the gold solutions. To allow for partial credit, we also measure the progress of the agents by checking if they reach specific `landmark' states in their solutions, such as completing a training stage. For the automatically generated problems, for which we have no gold solutions, we simply check if a key script (e.g., the training or evaluation script) was run successfully without exceptions, which we found to be an effective approximate metric.

We evaluate both proprietary and open-source LLMs on \mlee as the underlying models of strong baseline agents
with access to file-editing tools. We find that agents struggle to correctly solve many of the problems, with the strongest agent solving only 46.1\% of the \scenario sub-problems.
These agents are even further away from solving entire research tasks, completing correctly only 16.3\% of the end-to-end \emph{\expert} tasks. Open-source models substantially lag behind on both the sub-problems and end-to-end tasks.
Moreover, we find that the ranking of the agents and models on the \autogen set is mostly the same as it is on the curated sets, suggesting its potential usefulness for development.

Our analysis of model trajectories reveals that agents are better at resolving well-specified sub-problems, such as solving exceptions, bugs, and other issues, than tasks requiring repository and file exploration to understand code structure. 
These results underscore many of the core challenges facing LLM-based experiment execution systems, which our benchmark aims to help advance.

\section{Related Work}

\begin{table*}[!t]
    \centering
    \small
    \resizebox{\linewidth}{!}{
    \begin{tabular}{lccccc}
        \toprule
        \textbf{Resource} & \makecell{\textbf{\mlee}\\(this work)}& \makecell{ \textbf{DS-1000} \\\citep{ds1000} } & 
        \makecell{ \textbf{ML-Bench (Agent)}\\\citep{mlbench} } &  \makecell{ \textbf{MLAgentBench} \\\citep{mlagentbench} } & \makecell{ \textbf{SWE-bench}\\\citep{swebench} } 
        \\
         \cmidrule(lr){1-1}  \cmidrule(lr){2-2}  \cmidrule(lr){3-3}  \cmidrule(lr){4-4}  \cmidrule(lr){5-5} \cmidrule(lr){6-6}
          Repo. understanding   & \cmark & \xmark & \cmark & \xmark & \cmark  \\ 
          % Multi-File edits   & \cmark & \xmark & \xmark & \xmark & \cmark  \\ 
          % Partially-observed environment   & \cmark & \xmark & \xmark & \cmark & \xmark  \\ 
          Requires repository setup & \cmark & \xmark & \cmark & \xmark & \xmark  \\
          Outcome-based evaluation   & \cmark & \cmark & \xmark & \cmark & \cmark  \\ 
          \makecell[l]{Low-profile repositories\\ $\hookrightarrow$ Median stars}   & \makecell{\cmark\\14/14/23} & \makecell{\xmark\\35,309} & \makecell{\xmark\\9,632} & \makecell{-\\-} & \makecell{\xmark\\12,557}  \\ 
          \hdashline
          \# source repositories   & $\numtasks$/$\numtasks$/$\numautogen$ & $8$ & $18$ & - & $12$ \\ 
          \# problems   & $\numtasks$/$\numscenarios$/$\numautogen$ & $1000$ & $9641$ & $15$ & $2300$ \\ 
          % Number of gold LOC to execute   & $\mathbf{123}$ & $123$ & $123$ & $123$ & $123$ \\ 
         \bottomrule
    \end{tabular}
    }
    \caption{\small Comparison of \mlee against four other related code execution benchmarks in terms of the challenges being tested (rows 1-5) and the number of source repositories and problems in the dataset (row 6-7). For \mlee, number of repositories and problems refer to the \expert/\scenario/\autogen sets respectively. Repository understanding refers to the agent being required to go through repository files to complete a task. Repository setup refers to the requirement to install dependencies and environment. Outcome-based evaluation involves assessing performance based on unit-tests or comparing outcome results, such as metrics, to gold ones. Low-profile repositories refer to repositories with low number of GitHub stars.}
    \label{tab:dataset_comparison}
\end{table*}
% swe-bench & 12 & 27,844 (12,557) \\ 
%         DS1000 &  8 & 55,227 (35,309) \\ 
%         MLBench & 18 & 13,099 (9,632) \\ \hline 
%         \textbf{\mlee (\expert)} & \numtasks & 224 (14) \\
%         \textbf{\mlee (\autogen)} & \numautogen & 96 (23)  \\ 

\paragraph{Coding benchmarks:} While early code benchmarks \cite{humaneval, mbpp, multipl} mainly focused on synthesizing simple functions from descriptions, recent benchmarks have shifted to more complex competitive programming problems \cite{codecontests, apps, livecodebench} and evaluating proficiency with popular data science libraries \cite{ds1000}. Unlike these, we follow the recent trend on evaluating LLMs in more natural programming scenarios, such as programming with external tools and APIs \cite{apibank, taskbench, mint, patil2023gorilla}, code editing and debugging \cite{canitedit, DebugBench, FixEval}, resolving GitHub issues \cite{swebench} and understanding and coding within a repository context \cite{repobench, crosscodeeval, repoeval}.

In contrast to these works, \mlee focuses on the end-to-end task of setting up and executing research tasks in lower-profile repositories, presenting a unique set of challenges, with tasks that require repository comprehension and reasoning, editing multiple files, setting up the repository environment for execution while interactively running commands in the environment. Table~\ref{tab:dataset_comparison} compares the four datasets most relevant to \mlee. MLBench~\cite{mlbench}, specifically, its ML-Agent-Bench setup, evaluates LLMs' ability to execute tasks but focuses on popular code repositories rather than low profile, and does not evaluate based on correctness of outcome, i.e., whether resulting metrics are correct. MLAgentBench~\cite{mlagentbench} evaluates agents ability to run ML experiments but focuses on optimizing single-script ML experiments rather than comprehending and setting up arbitrary repositories for experimentation.

\paragraph{LLM Agents:} Recent advancements in LLM-based agents have shown significant progress across various domains, including games~\cite{voyager}, web navigation~\cite{webshop, webarena}, human interaction simulation~\cite{generativeagents}, automating complex computer tasks~\cite{osworld}, data science and machine learning~\cite{dsagent, datainterpreter, agenthpo, matplotagent}, open-ended discovery~\cite{discoveryworld}, and coding~\cite{codeact, sweagent, opendevin}. Our benchmark introduces an important new domain that encourages the development of LLM-based agents to assist researchers in their end to end research tasks with arbitrary repositories.

\section{Benchmark Construction}
\label{sec:benchmark}
\begin{figure}[!t]
    \centering
  \includegraphics[width=0.9\linewidth]{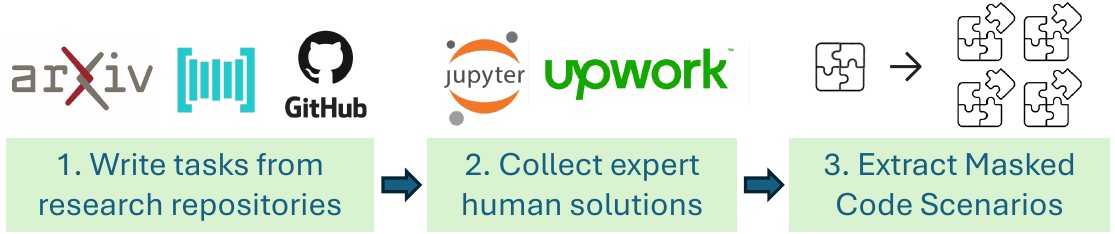}
    \caption{\small
    An overview of the construction pipeline for the \expert and \scenario sets. The Expert set contains manually written tasks, along with expert solutions (Step 2). The \scenario set contains problems extracted from the experts set (Step 3).
    }
    \label{fig:construct-overview}
\end{figure}

In this section we describe the process of building the \mlee benchmark.
The \mlee benchmark consists of 3 sets (see Table~\ref{tab:sets}) serving different purposes. The \textbf{\expert} set (\cref{subsec:expert}) contains manually written problems, solved by experts. The \textbf{\scenario} set (\cref{subsec:masked-problems}) contains sub-problems extracted from the \expert set using the gold solution, which provide easier and more focused sub-problems. \cref{fig:construct-overview} provides a high-level overview of the construction pipeline of these two sets. Finally, the \textbf{\autogen} set (\cref{subsec:autogen}) contains automatically generated problems which can be used for development and improvement of agents.

\paragraph{Environment Setup.}
Running research-oriented repositories often necessitates both being able to run system shell commands (e.g. to install dependencies and run scripts) and stateful Python commands. Previous work and environments typically support only one of these (e.g., only system shell commands \cite{swebench} or only Python commands \cite{mlagentbench}). Instead, we build an environment that allows running both of these commands with a Jupyter notebook as engine. In this setup, each execution code is equivalent to running a notebook \textit{cell}, which contains Python code and/or bash commands, and where state is reserved between cell executions (e.g., each cell can use any of the previously defined Python variables). The execution of each cell returns an observation string.

\subsection{\expert Set}
\label{subsec:expert}
We construct the \expert set by (1) identifying a set of relevant code repositories from research papers and manually writing research-oriented tasks based on them and (2) asking human experts to provide end-to-end solutions for these tasks (\cref{subsubsec:expert-construction}). 
We then use the expert solutions as the basis for \emph{outcome-based} evaluation, where we compare the agent's answer to the gold answer, and a more lenient \emph{landmark-based} evaluation that indicates progress toward correctly solving the task, even if the solution is not entirely correct (\cref{subsec:evaluation}).

\subsubsection{Construction}
\label{subsubsec:expert-construction}

\paragraph{Tasks.}
\label{subsec:tasks}
We create tasks motivated by the following two common settings: (1) reproducing numbers from research papers by running specific experiments, and (2) running modified experiments with different datasets, models, or configurations.

We start by collecting repositories from the ``Papers With Code'' (\url{github.com/paperswithcode/paperswithcode-data}) database, which contains research papers linked to their GitHub repositories, along with some additional metadata such as the modality of the datasets used. We only sample research papers with ``Text'' modalities and select repositories from 2021 or beyond.

We then manually review the sampled repositories and write tasks that involve running a single experiment that is mentioned either in the repository's ``readme'' file or under a script available in the repository, if such can be found.
Whenever possible, we make the task more challenging by requiring the experiment to be run on a new dataset or model, other than the one described in the available documentation. In these cases, we select either datasets available on HuggingFace Hub (\url{https://huggingface.co/datasets}) or provide a Google Drive link where the dataset can be found. The challenge of running on a specific dataset varies in difficulty: it could involve only a single configuration line change if the dataset is already supported, or creating a new dataset reader, adjusting column names, etc.

For each task, we define (1) the target Github repository, (2) the task definition (e.g., ``train a model...''), (3) the metrics or output to be reported (e.g., ``F1 metric on the validation set''), along with a specific structure of how the answer should be formatted, and (4) implementation instructions (e.g.~specific hyper-parameters).  The implementation instructions are important for two reasons: first, to allow fair evaluation of submitted answers by making sure the task is not under-specified, such that two agents that correctly complete the task get the same results. Second, to minimize computational requirements, as described next.

\paragraph{Minimizing Computational Requirements.}
To make \mlee faster and cheaper to run, we ensure tasks are executable without reliance on GPU machines and that they do not require more than 10 minutes of compute (e.g., for training models or installing packages) on basic compute instances (see \cref{sec:experiments} for compute details). We therefore create tasks that require minimal compute by only asking to train and evaluate small models (e.g., \texttt{gpt2-small}), and by adding implementation instructions to the task, such as ``only load the first 10 examples of the dataset'' or ``run a single epoch''. 

\begin{table}[!t]
    \centering
    \small
    \resizebox{\linewidth}{!}{
    \begin{tabular}{lcccc}
        \toprule
        Set & \# & Solutions & Evaluation & Purpose \\
\midrule
\expert & \numtasks & \cmark & Solution-based & Benchmark \\
$\hookrightarrow$ \scenario & \numscenarios & \cmark & Solution-based & Benchmark, analysis \\
\autogen & \numautogen & \xmark & Proxy & Development \\
        \bottomrule
    \end{tabular}
    }
    \caption{The different sets of \mlee.}
    \label{tab:sets}
\end{table}

Note that these restrictions do not make the task any easier for the agent. In fact, they often add additional challenges that agents need to solve (e.g., configuring hyper-parameters, finding where data is loaded to limit its loading to the first 10 samples, being able to run experiments that were designed for GPU on a CPU, etc.).

\paragraph{Expert Annotation.}
\label{subsec:experts}
We use Upwork (\url{https://www.upwork.com/}) to find and hire experts that have experience with running ML and NLP experiments. We filter the initial list of applications by testing them on a pilot task which we solved ourselves, to make sure they are able to correctly execute a task and effectively solve issues by comparing their solution and results to ours. We instruct workers to execute their solutions on Google Colab, allowing us to collect the solutions in a consistent notebook-like environment.

We ask the experts to submit their (1) solution notebook, (2) answers (i.e. metrics to be reported), the specific git commit hash of the repository that they have used, and the final version list of all dependencies that were installed throughout the notebook.\footnote{The git hash and dependencies ensure that these solutions can be reproduced in the future even as repository and package versions change.} In addition, we instruct them to use default parameters whenever possible, and to report any decision that they had to make but was not specified in the original task (e.g., the selection of a delimiter token or specific hyper-parameters when no default values are provided). We add these decisions to the task description or implementation instructions to ensure that any agent solving the same task would have all the necessary information needed to get the same results.

Finally, we manually review their solutions, making sure that (1) the solution correctly follows the task, (2) it can be executed in our environment, (3) all unspecified decisions have been recorded and (4) re-running the experiment multiple times yields the same results (up to an error of $10^{-2}$). If needed, we ask the workers to make corrections, or manually fix issues ourselves.
Solutions that we could not execute on our Jupyter environment, such as solutions that had to modify the installed Python version were discarded.
We provide cost details and guidelines in \cref{app:experts}.

\subsubsection{Evaluation}
\label{subsec:evaluation}

\paragraph{Accuracy Evaluation.} As described in~\S\ref{subsec:tasks}, experts provide us a deterministic solution for each task, which we then execute in our environment to get the gold answer, allowing us to evaluate agents based on their outcome. Answers consist of several values (e.g., numbers for metrics, string for model predictions). We define the accuracy metric as the portion of correctly answered values: where the predicted answer precisely matches the gold one (up to a $10^{-2}$ error). Unlike reference based evaluation used in cloze tests and various prior coding benchmarks~\cite{repobench,mlbench}, outcome-based evaluation 
% does not require code to match the gold cells and
allows for alternate valid solutions. 

\paragraph{Landmark-Based Evaluation.} Sometimes an indication of whether the model was precisely correct may be too strict, ``punishing'' models that make progress but don't reach the end. E.g., an agent that loads the data but doesn't train would have the same accuracy as an agent that fails at the start.

To measure progress towards the final goal, we use the gold task notebooks to identify \textit{landmark outputs}; outputs from the environments that act as ``evidence'' that a particular step was run successfully. E.g., the explicit output string ``\texttt{*** training completed ***}'' in Figure~\ref{fig:example} or the string ``\texttt{Loading data... 100\%}'' implying successful data loading.

Importantly, a perfect landmark score does not entail a perfect accuracy score, as landmarks only indicate that some action was performed, but it was not necessarily correct (e.g., a training script run successfully but with wrong hyper-parameters could be counted as success). Similarly, albeit unlikely by design, a model could correctly solve the task but not hit all of the landmarks (e.g., if it uses an alternate approach or guesses a solution) and have a lower landmark score.
For each gold solution we manually extract 2-6 landmark outputs patterns. The landmarks metric evaluates the percentage of these patterns that appear in the outputs of any of the cells executed by the agent.

\subsection{Masked Coding sub-problems Extraction}
\label{subsec:masked-problems}

Solving an end-to-end execution task can often be long and complex, consisting of multiple non-trivial steps. As such, evaluating agents on their ability to run an entire task provides a sparse signal of success, where agents have to complete numerous steps correctly to succeed, making it harder to ``hill-climb'' results. Instead, we want to evaluate models in a more fine-grained way that will allow us to get success signals for any incremental progress towards the task. To this end, we propose to focus on a specific sub-problem from the task solution at a time, leveraging the expert solutions from the expert set.

\begin{figure}[!t]
    \centering
  \includegraphics[width=0.95\linewidth]{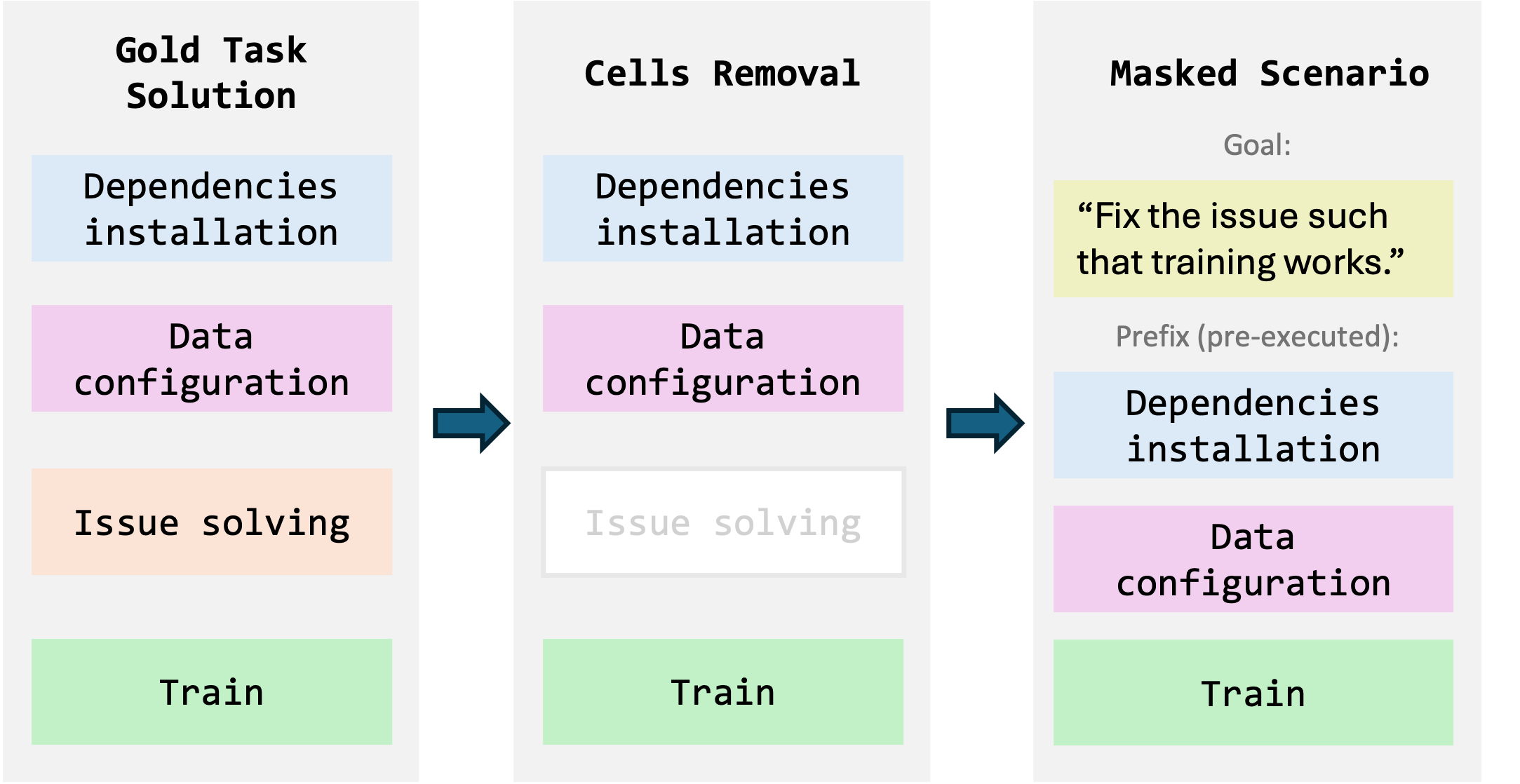}
    \caption{\small
    An abstract demonstration of how sub-problems are extracted: starting from a gold end-to-end task solution (left), we remove cells (middle) that focus on certain aspects (differently colored cells in the figure), then create a masked problem by defining a goal and prefix cells (right). The prefix cells are executed in the environment, and the agent must then write code to solve the sub-problem.
    }
    \label{fig:problem-extraction}
\end{figure}

We turn to an approach loosely inspired by cloze tests \cite{taylor1953cloze} and masked language models (MLM; \citealp{devlin-etal-2019-bert}): given a gold solution for a task we remove (\textit{mask}) some part of it (e.g., code that solves a dependency installation issue), and manually define the sub-problem such that an agent only needs to solve this narrower aspect of the overall task, as captured by the removed cells.

While the ideal goal is for the agents to complete tasks in an end-to-end manner, extracting masked sub-problems allows us to evaluate agents on a broad range of technical challenges often encountered when working with research repositories, while performing a finer-grained analysis of their performance. In addition, this setup aligns well with the usage of interactive code assistants (e.g., CoPilot and Colab's AI-powered coding), where agents assist users that have already written or even executed some code, and can specify what issue or problem remains to be solved.

\paragraph{Masked Coding Sub-Problems.}
Each masked sub-problem consists of (1) a textual description of the remaining tasks to execute or issues to be resolved (e.g., ``fix the runtime error to complete training'' for a sub-problem where cells that fix a runtime error were removed), and (2) code prefix (e.g., the other cells from the original notebook that were not masked). Given a masked sub-problem, the code prefix is pre-executed in the environment, and agents must then execute their own cells to solve the sub-problem.

\paragraph{Extraction Procedure.} We extract masked coding sub-problems in a manual process where we first identify certain cells (not necessarily consecutive) in the gold solution that focus on a certain aspect. For example, the cells corresponding to loading and modifying the dataset could be combined into a \textit{data configuration} block as shown in pink  in Fig.~\ref{fig:problem-extraction}. We extract sub-problems by first masking a block (\textit{Issue solving} orange block in the figure). We then identify cells that do not depend on the masked block and define a goal that is remaining to be completed (e.g., making training run in our example). These cells are pre-executed and the agent's task is to complete the goal.

We use the masked block and goal cell to define the sub-problem, e.g., if the code to handle execution on CPU has been masked, the sub-problem definition would be ``\textit{I have already executed some of the required steps. Now, you should make the necessary changes to make sure the code runs on a CPU. Your goal is to successfully run `train.py'.}''. We choose cells with clearly identifiable success indicators as goals, e.g., successfully running `train.py' would produce metrics on completion.

\paragraph{Evaluation.}
Since each sub-problem has a clearly defined goal, extracted from the original task, we can use the same outputs and landmarks as in the expert set, and similarly evaluate accuracy and landmarks (\cref{subsec:evaluation}).
We evaluate sub-problems with the same metrics defined in \cref{subsec:evaluation}.

\begin{table*}[!t]
    \centering
    \footnotesize
    \begin{tabular}{lrr p{0.6\linewidth}}
            \toprule
            Category (\%) & Portion & Gold LOC & Example(s) description of a gold solution \\ 
            \midrule
            Dependencies & 19.7\% & 4.1 & Downgrade `transformers` version to allow  execution of an older repository \\
            CPU & 7.2\% & 5.1 & Remove `\texttt{.cuda()}` from different locations in code \\ 
            Configuration & 12.5\% & 8.2 & Edit Python or shell scripts to set hyper-parameters and experiment details\\
            Data & 23.7\% & 22.7 & Download custom dataset, update data loader, limit to loading first 10 samples\\
            Issue & 9.2\% & 5.8 & Pytorch incompatible tensor shapes; incorrectly loaded Python package\\
            Goal & 25.0\% & 6.5 & Run the evaluation script then load generated file to report metrics \\
            Other & 2.6\% & 3.8 & Save the model after training to allow evaluation loading \\
            \bottomrule
        \end{tabular}
    \caption{Distribution of sub problems categories and description of representative solutions from experts. LOC stands for lines of code, counting the number of lines (excluding comments) in the gold solution. }
    \label{tab:sub-problems-examples}
\end{table*}

\subsection{Automatically Generated Tasks}
\label{subsec:autogen}
The \expert and \scenario sets provide validated problems and reproducible solutions, allowing for more accurate evaluation of agents. However, creating expert tasks is both time-consuming and costly, and the limited number of tasks hinders their use for agent improvements, such as fine-tuning models based on trajectories and environment feedback \cite[e.g.,][]{Chen2023FireActTL,song-etal-2024-trial,yin-etal-2024-agent}. To address this, we automatically generate tasks using an LLM (namely, GPT-4o).

\subsubsection{Construction}
Generation involves two steps: (1) filtering suitable repositories, and (2) generating tasks based on the readmes of these repositories. 

\paragraph{Filtering.} We start with the same list of repositories from \cref{subsec:tasks}, selecting those listed in 2021 or later, resulting in a total of 5915 repositories. Many of these repositories cannot be trivially used to generate tasks due to various limitations: some do not support running any experiment, some have missing dataset links, some require GPU hardware, and some depend on APIs of other LLMs (such as OpenAI or Anthropic), which could be unavailable or incur costs for users. To alleviate these issues, we employ a combination of heuristic and LLM-based filtering methods. Specifically, we keep repositories that mention specific datasets and models, do not use APIs, and do not require GPUs. Detailed information on the filters and the prompt used for LLM-based filtering can be found in \cref{app:autogen}.

\paragraph{Creating Tasks.} Given the filtered repositories, we prompt the LLM with the contents of each repository's README file and instruct it to create an experiment-running task. This includes defining the goal, dataset, model, and script to be run (\textit{``Run probability-based prompt selection on the SST-2 dataset using opt-125m as the base model with the script `run\_prompt\_selection.py`''}). We also specify that the LLM should choose the smallest model within model families (e.g., BERT-base if BERT models are supported). See \cref{app:autogen} for further details on the generation process. To verify the quality of the generated set, we sample 100 generated tasks and find that 81\% of the samples are feasible; among the rest, the most prominent issue was missing resources (dead or gated links to datasets, missing code) and conceptually flawed tasks such as asking to use a discriminative model for a generative task. 

\paragraph{Difference from Expert Set.} Importantly, the \autogen tasks exhibit different properties than those of the \expert set: problems in the \expert set can require training or inference on datasets not specifically mentioned to be supported, in some cases with different formats and columns, whereas \autogen tasks focus more on getting the dependencies installed and being able to start an experiment. Moreover, the \expert set only includes problems where we were able to run the solutions in our Jupyter environment, while problems in the \autogen set could potentially involve even more challenging environments setups we avoided, such as when changing a Python version is required. Finally, \autogen tasks sometimes require navigating through web pages that were mentioned in the repository's Readme file to download datasets.

\subsubsection{Evaluation} 
\label{subsec:autogen:evaluation}
Without expert solutions, we cannot evaluate based on outcomes or landmarks. Instead, we use a simple heuristic metric, termed \textit{Script-Executed}, to ensure the model sets up and executes the experiment without unresolved issues: we check if the script the agent was asked to run executes without exceptions for a minimum duration (see \Cref{app:autogen:eval} for details). The minimum duration ensures that the script was successful and didn't just fail silently. While this method does not guarantee perfect evaluation, we find it surprisingly effective, as we show in our analysis in \Cref{subsec:error_analysis}.

\subsection{\mlee Benchmark}
\label{subsec:data-analysis}

The \expert set consists of \numtasks collected tasks, where each problem is paired with a gold output for outcome-based evaluation and a small set of landmarks for our softer evaluation metric (an average of 3 landmarks per problem).

To roughly estimate how much lines of code (LOC) are needed to solve a task, we count the number of lines (excluding comments) in the gold solution, and for editing cells, the number of changed lines.\footnote{The gold LOC can be dramatically lower than those needed by the agent; experts did not have to write code to read contents of files in the repository as they can use their IDE or browser,  while agents have to browse files through an explicit command. In addition, experts did not necessarily keep cells with failed attempts.}
An average of 44.3 LOC and 14.4 cells per solution suggest that these tasks are particularly challenging due to potentially long agent trajectories. Consequently, the ability to solve these tasks provides an important signal on the performance of agents in handling long and complex tasks.

The \scenario set contains \numscenarios masked coding sub-problems derived from the \numtasks expert tasks. Like the \expert set, each sub-problem is paired with a gold output and landmarks. \cref{tab:sub-problems-examples} shows the distribution of the extracted sub-problems across various categories, along with a representative solution for each category and the average lines of code (LOC) that were executed in the gold solution. Finally, our automated set includes \numautogen problems, all from unique repositories.

To verify that the repositories used in our benchmarks are indeed `low-profile', we count the number of GitHub stars in the source repositories as a proxy for popularity as shown in Table~\ref{tab:dataset_comparison}. Intuitively, popularity loosely correlates with the quality of documentation and readiness, which affects the difficulty of experiments execution. We see that the median number of stars for our repositories (14) is considerably lower than other comparable datasets (see Appendix~\ref{app:repo_details} for details).
\section{Experiments}
\label{sec:experiments}

\paragraph{Experimental Setup.}
We limit the execution time (i.e. time to run commands, not counting the API reply time) of each problem to 30 minutes. We run all tasks on compute instances using sandboxed execution in Modal (\url{https://modal.com}), which allows us to evaluate problems safely and concurrently, speeding up evaluations. We limit the total input tokens number (summing all steps) of each sub-problem to 400k, and of expert and auto-generated tasks to 600k. The compute cost for executing each problem in Modal (not counting API usage) is 2-3 cents, making it negligible compared to API costs. If time or token limit occurs before the agent submits the answer, the task is concluded without submission (in such cases, agents get 0 accuracy, but are still scored based on the landmark evaluation). We validate that running the gold trajectories yields perfect scores on all of our metrics in three consequent attempts. To estimate variance in our results, we average results over 3 seeds for the expert set tasks with the strongest underlying LLM (to limit costs).

\paragraph{Underlying LLMs.}

We experiment with agents based on commerical LLMs GPT-4o (\texttt{gpt-4o-2024-08-06}) and GPT-4o mini (\texttt{gpt-4o-mini-2024-07-18}) \cite{openai2023gpt4}, as well as the open-source models Mixtral-8x22B-Instruct \cite{jiang2024mixtral} and Llama 3.1 70B \cite{dubey2024llama3herdmodels} (\texttt{Meta-Llama-3.1-70B-Instruct-Turbo}), both served by \url{https://www.together.ai/}.

\subsection{Baselines}
\label{subsec:baselines}

In this section, we describe the three baseline agents that we evaluate on {\mlee}: ReAct \citep{yao2023react}, our improved version \mlereact, and SWE-Agent \cite{sweagent}. All of our agents are given access to a Jupyter notebook environment where they can \textbf{execute} Python or Bash commands. Other than the execute action, they can also \textbf{submit} an answer when done. For sub-problems, we execute the provided `pre-execute' cells (\S\ref{subsec:masked-problems}) and pass them as an existing history of actions to each agent. With end-to-end tasks (\expert and \autogen), we simply prompt the agent with the task description.

\paragraph{ReAct}~\cite{yao2023react} is a baseline agent that iteratively prompts the underlying LLM to output both an action and a natural language ``thought'', providing the interaction history as context. Each step, the generated action is executed against the environment and a $<$\texttt{thought}, \texttt{action}, \texttt{observation}$>$ tuple is added to the history, until the agent submits an answer, or exceeds token or compute limitations.

One challenge associated with running experiments is that output observations can get extremely long (e.g., the output of certain training scripts and dependency installation reach 10k-40k tokens).
ReAct agents ``accumulate'' history information (thought, action, and observation triplets) at each step, which makes token usage grow rapidly. As agents are typically limited to a fixed budget (either cost or tokens), this could lead to failures. We apply truncation strategies in all our agents and baselines to mitigate this issue. See \cref{app:prompts} for details.

\paragraph{\mlereact.}
The ability to execute Python and bash commands, in theory, allows agents to perform any necessary task. However, these actions are still limited compared to humans who can use IDEs to browse and edit files. In our early experiments, we indeed found that agents struggle to edit files (e.g., change configs) using just bash. 

To address this challenge, we supplement the agent with an additional \textbf{edit} action, similar in spirit to the \emph{Agent-Computer Interfaces} from SWE-Agent~\cite{sweagent}. Specifically, the edit command accepts three parameters: the name of the file, the exact content of the lines to be replaced, and the content to replace it with. We do not ask the agent to provide line numbers (as needed by git patches or SWE-Agent), and provide the agent with suggestions in case the exact content of lines to be replaced were not found (e.g., if whitespaces are missing). See App.~\ref{app:edit} for more details. 

\begin{table}[!t]
    \setlength{\tabcolsep}{1ex}
    \centering
    \footnotesize
    \begin{tabular}{llll}
        \toprule
        \textbf{Agent} & \textbf{Model} & \textbf{Acc.} & \textbf{Landm.}\\ 
        \midrule
SWE-Agent & GPT-4o & 16.3 $\pm$ 2.1 & 36.8 $\pm$ 2.3 \\ 
React & GPT-4o & 12.2 $\pm$ 1.0 & 33.6 $\pm$ 0.9 \\ 
React-Super & GPT-4o & 14.4 $\pm$ 2.2 & 42.6 $\pm$ 2.9 \\ 
\hdashline
SWE-Agent & GPT-4o mini & \ \ 3.3 & 16.1 \\ 
React-Super & GPT-4o mini & \ \ 5.6 & 20.6 \\ 
\hdashline
SWE-Agent & Llama 3.1 70B & \ \ 5.6 & \ \ 4.8 \\ 
React-Super & Llama 3.1 70B & \ \ 6.1 & \ \ 9.6 \\ 
\hdashline
SWE-Agent & Mixtral 8x22B & \ \ 1.1 & \ \ 0.0 \\ 
React-Super & Mixtral 8x22B & \ \ 3.3 & \ \ 3.7 \\ 
        \bottomrule
    \end{tabular}
    \caption{Results on \expert,
    % with different underlying LLMs,
    with GPT-4o numbers averaged across 3 seeds.}
    \label{tab:results-e2e}
\end{table}

\paragraph{SWE-Agent} \citep{sweagent} is a ReAct-based agent, originally designed to solve GitHub \textit{issues}. Like \mlereact, this approach provides agents with tools that allow easier editing of files, but also tools for reading files, scrolling through their contents and more (see original paper for details). We implement this agent in our environment and modify the prompt to address the execution of research tasks in our environment.

\paragraph{Reflecting Agents.}
To explore whether agents can improve their performance by reflecting on their failures, we evaluate agents with a reflection mechanism \cite{reflexion}. Whenever an agent's first attempt to complete a task fails to submit any answer, we prompt the underlying LLM to reflect on the trajectory and devise a plan to avoid the same mistakes in subsequent attempts. This reflection is then incorporated into the agent's prompt and it tries again to solve the problem. The agent is given $k$ tries to solve the problem with each try being given $1/k^{th}$ of the token budget.

\subsection{Results}

\paragraph{\expert Set.} We show results for the experts set in Table~\ref{tab:results-e2e}, with the results for the most performant LLM averaged across three seeds (decoding temperature is 0.2). The low accuracies (12.2-16.3) suggest that current agents cannot yet perform this task well. However, in some cases, agents make some progress towards the goal, as evident by the landmarks metric, suggesting that agents could still be helpful in setting up repositories.

\paragraph{\scenario Set.} We show results in Table~\ref{tab:results-problems} on the \scenario set, demonstrating that SWE-agent correctly solves a significant portion (46.1\%)
of the challenges that are required to set-up and execute experiments from research repositories, but that most sub-problems are still unsolved. The higher landmarks evaluation score (74.9\%) suggests that agents often make progress towards solving the sub-problems, even if some of the steps might not necessarily be correct.

We find that SWE-agent performs better than \mlereact with GPT-4o as the LLM, but slightly worse with all weaker models, suggesting that weaker models were less effectively in leveraging SWE-Agent tools. The open-source Mixtral and Llama  reach significantly lower scores on both the \scenario and \expert sets.

\begin{table}[!t]
    \centering
    \footnotesize
    \begin{tabular}{llll}
        \toprule
        \textbf{Agent} & \textbf{Model} & \textbf{Script-Executed}\\ 
        \midrule
        SWE-Agent & GPT-4o & 18.0 \\ 
React & GPT-4o & 14.0 \\ 
React-Super & GPT-4o & 18.8 \\ 
\hdashline
SWE-Agent & GPT-4o mini & \ \ 5.2 \\ 
React & GPT-4o mini & 16.0 \\ 
React-Super & GPT-4o mini & 14.8 \\ 
        \bottomrule
    \end{tabular}
    \caption{Results on 250 of the \autogen tasks.}
    \label{tab:results-autogen}
\end{table}

\paragraph{\autogen Set.}
We show in Table~\ref{tab:results-autogen} results for the \autogen tasks, where ranking of models and agents are mostly consistent with the ranking of the models on the \scenario set, suggesting potential usefulness of this set for future development.

\paragraph{Ablations.}
Comparing React with \mlereact shows that the editing function enables the agent to hit more landmarks (72.5\% vs 65.7\%) and produce more accurate answers (41.6\% vs 37.0\%). We find that without the editing command, the agent usually resorts to editing files with the \texttt{sed} command, which is designed for simple single-line edits.

\paragraph{Can agents that reflect do better?}
We next evaluate if retrying after reflecting on failures can improve the performance of our baseline agents, with $k=3$ retries. As shown in Table~\ref{tab:results-reflect}, the additional retries with reflections have a positive but minimal impact on the score. If models lack the inherent ability to resolve some of these issues, retrial with reflections are not likely to help.

% \begin{table}[!t]
%     \centering
%     \footnotesize
%     \begin{tabular}{llll}
%         \toprule
%         \textbf{LLM / Agent} & \textbf{Subm.}  &  \textbf{Acc.} & \textbf{Landm.}\\ 
%         \midrule
%         GPT-4o & & & \\
%         \; SWE-agent    & 33.6 & 23.7 & 42.6 \\
%         \; ReAct & 50.0 & 32.9 & 56.7 \\
%         \; \mlereact & 55.3 & 34.9 & 64.7 \\
%         \hdashline
%         GPT-4-turbo & & & \\
%         \; SWE-agent & 28.9 & 18.4 &  34.8 \\
%         \; \mlereact & 51.3  &  33.6  & 57.0 \\
%         \hdashline
%         Mixtral & & & \\
%         \; SWE-agent & 0.0 & 0.0 & 0.0  \\
%         \; \mlereact & 0.0 & 0.0 & 0.0  \\
%         \bottomrule
%     \end{tabular}
%     % \caption{Results on 150 problems\bb{placeholder..}, fixed price limit per task. \bb{temporary names of agents}}
%     \caption{Results of our baselines on \mlee with different underlying LLMs.}
%     \label{tab:results-problems}
% \end{table}

\begin{table}[!t]
    \centering
    \footnotesize
    \begin{tabular}{llll}
        \toprule
        \textbf{Agent} & \textbf{Model} & \textbf{Acc.} & \textbf{Landm.}\\ 
        \midrule
SWE-Agent & GPT-4o & 46.1 & 74.9 \\ 
React & GPT-4o & 37.0 & 65.7 \\ 
React-Super & GPT-4o & 41.6 & 72.5 \\ 
\hdashline
SWE-Agent & GPT-4o mini & 27.0 & 51.8 \\ 
React-Super & GPT-4o mini & 31.5 & 58.3 \\ 
\hdashline
SWE-Agent & Llama 3.1 70B & 17.4 & 35.0 \\ 
React-Super & Llama 3.1 70B & 22.8 & 38.3 \\ 
\hdashline
SWE-Agent & Mixtral 8x22B & \ \ 9.5 & 26.6 \\ 
React-Super & Mixtral 8x22B & \ \ 7.0 & 13.2 \\ 
        \bottomrule
    \end{tabular}
    % \caption{Results on 150 problems\bb{placeholder..}, fixed price limit per task. \bb{temporary names of agents}}
    \caption{Results of our baselines on \mlee (\scenario) with different underlying LLMs.}
    \label{tab:results-problems}
    \vspace{3mm}
\end{table}

\subsection{Error Analysis}
\label{subsec:error_analysis}
The \scenario set categorizes each problem, allowing us to break down performance of agents (\cref{tab:sub-problems-examples}). We find that the hardest categories for the agent are data (27\%), configuration (38\%) and goal (43\% accuracy), whereas CPU, issues and dependencies are easier (73\%, 61\% and 54\% respectively).
These findings suggest that agents are better at solving sub-problems where there is a specific error message to be solved (such as CPU support errors, incompatible dependencies, or exceptions) than more open-ended problems such as configuring data loading for a custom dataset.

Specifically, for the latter case, we find that agents commonly skip going through the repository to understand relevant code. For example, they often hallucinate arguments of scripts or functions instead of looking up how they should be called (e.g., adding \texttt{n\_examples=10} when no such argument is defined), or the opposite: they 
miss a script parameter and attempt to change them in files unsuccessfully. Additionally, once they commit to a particular approach, they never reconsider their decision until failure.
These issues suggest that agents should be better designed to analyze repositories and consider multiple solution approaches.
We provide all trajectories of \mlereact and SWE-Agent in our code repository.\footnote{\url{https://github.com/allenai/super-benchmark/tree/main/trajectories}}

% \begin{table}[!t]
%     \centering
%     \footnotesize
%     \begin{tabular}{lccc}
%         \toprule
%         \textbf{Agent} & \textbf{Subm.} & \textbf{Acc.} & \textbf{Landm.}\\
%         \midrule
%         \mlereact & 55.3 & 34.9 & 64.7 \\
%         Retry &  &  &  \\
%         Reflexion &  53.3 & 31.6 & 57.8 \\
%         \midrule
% %         \cline{2-4} \\
%          \textbf{2x budget} & & & \\
%         \;\mlereact & 59.2 & 37.5 & 66.1 \\
%         \;Retry &  &  &  \\
%         \;Reflexion &  53.3 & 31.6 & 57.8 \\
%         \bottomrule
%     \end{tabular}
%     \caption{\tbd{caption}}
%     \label{tab:results-reflect}
% \end{table}

\begin{table}[!t]
    \centering
    \footnotesize
    \begin{tabular}{lccc}
        \toprule
        \textbf{Agent} & \textbf{Acc.} & \textbf{Landm.}\\
        \midrule
        \mlereact & 41.6 & 72.5 \\
        %Retry &  &  62.9 & 66.0 \\
        Reflexion~\cite{reflexion} & 45.4 & 76.6 \\
        % \midrule
        % \textbf{2x budget} & & & \\
        % \;\mlereact & 37.7 & 59.6 & 66.5 \\
        % %\;Retry &  &  &  \\
        % \;Reflexion & 31.6 & 53.3 & 57.8 \\
        \bottomrule
    \end{tabular}
    \caption{\small Results of the \mlereact agent (using GPT-4o) with and without Reflexion on the \scenario set. While retrying with reflection does help improve the submission rate and accuracy, \mlee benchmark still remains challenging.}
    \label{tab:results-reflect}
    \vspace{2pt}
\end{table}

\paragraph{Effectiveness of proxy metric.} While we evaluate the \expert and \scenario sets based on solutions of experts, for the \autogen set we have no such solutions, and therefore rely on the weaker Script-Executed proxy metric (\Cref{subsec:autogen:evaluation}). To verify that this proxy metric is reliable, we use the trajectories of \mlereact on the \scenario set to compare the Script-Executed metric with the accuracy and landmark metrics. We find that Script-Executed agrees with landmark (assuming a score of 1 when landmark > 0.5) in 90\% of the cases and with the accuracy metric in 69\% of the cases. We identified two cases of disagreement with the landmark metric: (1) the target script ran sufficiently to get the correct answer, but still encountered an exception at the end (e.g., an exception creating a figure) which would be considered incorrect by the proxy metric (2) the script ran for the minimum time making it appear like a success based on the proxy metric only to fail due to a mis-configuration or exceptions much later (and not reaching the answer).

\section{Conclusion}
Our work introduces \mlee, a benchmark designed to evaluate LLM-based agents on executing tasks from code repositories, focusing specifically on low-profile research repositories encountered in the wild.  We show empirically that our benchmark is difficult, even for the current best commercial LLMs such as GPT4,  both on landmark and end-to-end task evaluations (e.g., GPT-4o solving only 46.1\% of the sub-problems). Our benchmark also highlights many of the core challenges in building autonomous LLM-based execution agents, such as repository reasoning and code editing, which we hope will help the community make measurable progress on this important problem.

\section*{Limitations}

\paragraph{Dataset Size.} The dataset size of our benchmark, comprising \numtasks and \numscenarios sub-problems, is smaller compared to some other benchmarks available for agent evaluation, which could potentially affect the statistical significance of performance evaluations. However, the use of smaller, high-quality benchmarks is not uncommon. For instance, benchmarks such as \textsc{HumanEval} \cite{humaneval}, \textsc{ClassEval} \cite{classeval}, and \textsc{Bamboogle} \cite{press-etal-2023-measuring} contain 164, 100, and 125 examples respectively, and are widely used for assessing model performance. In addition, recent work has suggested that reducing large datasets to as few as 100 examples does not diminish their effectiveness \cite{polo2024tinybenchmarks}.
Moreover, smaller-sized datasets offer the advantage of being less expensive to operate, thus providing better accessibility for researchers with limited resources, particularly when running interactive agents in environments that generate long outputs. Finally, our provided \autogen set with \numautogen problems offers problems purposed for development, which alleviates the risk of overfitting to the evaluation sets.

\paragraph{Programming Languages and Domains.} We have only collected solutions written in Python, and our environment only supports that programming language. We focus mostly on text-based repositories. While the challenges associated with running these repositories likely overlap with other domains, increasing the diversity of the repository domains could be beneficial.

\paragraph{Evaluation Based on External Resources.} Running benchmarks in realistic environments often depend on external resources. In our case, agents rely on availability of resources such as GitHub, pip and datasets, which we cannot control across runs. While completely sand-boxed setups could have allowed for a more controlled evaluation, we opt for fidelity, similarly to e.g. benchmarks for web agents that rely on access to real websites \cite[][inter alia]{mialon2024gaia,he2024webvoyager}.\footnote{Note that all our evaluations are run using the same base Docker image with sand-boxed code execution and should be reproducible; barring any external changes.}

\section*{Ethical Considerations}

While autonomous research execution agents could significantly enhance research advancements, there is a risk of over-reliance on these agents, which could lead to conclusions drawn based on incorrect implementations of agents, and careless actors not checking the agent's reproduction work carefully.

\section*{Acknowledgments}
We thank Ori Yoran for his valuable comments and suggestions. We also thank the Upworker expert programmers for their work on the solutions to the Expert set problems.

\bibliography{anthology,SUPER}

\clearpage
\appendix

\onecolumn

\section{Repository details}
\label{app:repo_details}

Table~\ref{tab:repo_info} shows information about the \numtasks source repositories used to create the \expert and \scenario sets, including their name, original GitHub link and the number of stars on GitHub.

\begin{table*}[h]
\centering 

{\scriptsize
 \begin{tabular}{ | p{3cm} p{10cm} c  | }
        \hline 
          \textbf{Task} & \textbf{GitHub} & \textbf{Stars}  \\ \hline 
              colbert & \url{ https://github.com/stanford-futuredata/ColBERT } & 2826\\
	textbox & \url{ https://github.com/RUCAIBox/TextBox } & 1069\\
	amrbart & \url{ https://github.com/goodbai-nlp/AMRBART } & 94\\
	g-transformer & \url{ https://github.com/baoguangsheng/g-transformer } & 43\\
	pie-perf & \url{ https://github.com/madaan/pie-perf } & 80\\
	safetybench & \url{ https://github.com/thu-coai/SafetyBench } & 138\\
	discodisco & \url{ https://github.com/gucorpling/DisCoDisCo } & 6\\
	acqsurvey & \url{ https://github.com/rahmanidashti/acqsurvey } & 11\\
	curriculum\_learning & \url{ https://github.com/adymaharana/curriculum_learning } & 9\\
	spa & \url{ https://github.com/OceannTwT/SPA } & 5\\
	mezo & \url{ https://github.com/princeton-nlp/MeZO } & 1016\\
	mode-connectivity-plm & \url{ https://github.com/thunlp/mode-connectivity-plm } & 7\\
	mbib & \url{ https://github.com/Media-Bias-Group/MBIB } & 22\\
	quantifying-stereotypes-in-... & \url{ https://github.com/nlply/quantifying-stereotypes-in-language } & 1\\
	rah-kbqa & \url{ https://github.com/yanmenxue/rah-kbqa } & 6\\
	dir-gnn & \url{ https://github.com/wuyxin/dir-gnn } & 115\\
	unsupervisedhierarchicalsymbolic... & \url{ https://github.com/SiyuLou/UnsupervisedHierarchicalSymbolicRegression } & 0\\
	conv\_graph & \url{ https://github.com/huawei-noah/noah-research/tree/master/conv_graph } & 0\\
	mera & \url{ https://github.com/ai-forever/MERA } & 55\\
	pira & \url{ https://github.com/C4AI/Pira } & 5\\
	pet & \url{ https://github.com/timoschick/pet } & 1618\\
	transnormerllm & \url{ https://github.com/opennlplab/transnormerllm } & 221\\
	bert-lnl & \url{ https://github.com/uds-lsv/BERT-LNL } & 9\\
	blockskim & \url{ https://github.com/chandlerguan/blockskim } & 6\\
	data\_label\_alignment & \url{ https://github.com/gyauney/data-label-alignment } & 3\\
	hype & \url{ https://github.com/yuanhy1997/HyPe } & 13\\
	paraphrase-nli & \url{ https://github.com/matejklemen/paraphrase-nli } & 3\\
	powerfulpromptft & \url{ https://github.com/zhengxiangshi/powerfulpromptft } & 71\\
	robust\_prompt\_classifier & \url{ https://github.com/adianliusie/robust-prompt-classifier } & 5\\
	align-to-distill & \url{ https://github.com/ncsoft/Align-to-Distill } & 4\\
	inbedder & \url{ https://github.com/zhang-yu-wei/InBedder } & 20\\
	transpolymer & \url{ https://github.com/ChangwenXu98/TransPolymer } & 51\\
	memorizing-transformers-... & \url{ https://github.com/lucidrains/memorizing-transformers-pytorch } & 622\\
	multi3woz & \url{ https://github.com/cambridgeltl/multi3woz } & 14\\
	galore & \url{ https://github.com/jiaweizzhao/galore } & 1332\\
	amos & \url{ https://github.com/microsoft/amos } & 24\\
	glee & \url{ https://github.com/genezc/Glee } & 9\\
	parallel-context-windows & \url{ https://github.com/AI21Labs/Parallel-Context-Windows } & 98\\
	logme-nlp & \url{ https://github.com/mainlp/logme-nlp } & 5\\
	mixup-amp & \url{ https://github.com/pai-smallisallyourneed/mixup-amp } & 4\\
	upet & \url{ https://github.com/wjn1996/UPET } & 2\\
	dpt & \url{ https://github.com/xyaoooo/dpt } & 6\\
	team & \url{ https://github.com/declare-lab/team } & 22\\
	cet & \url{ https://github.com/zzz47zzz/CET } & 18\\
	linkbert & \url{ https://github.com/michiyasunaga/LinkBERT } & 411
        \\ \hline  
 \end{tabular}
}

\caption{Details of the \numtasks repositories used in {\mlee} along with GitHub link and star information as of September 3rd, 2024.}
\label{tab:repo_info}
\end{table*}

Adding to the information in Table~\ref{tab:dataset_comparison}, we show below the average and median star ratings for other comparable benchmarks (all star ratings are collected as of \texttt{September 3rd, 2024}). This was computed automatically from the GitHub API based on the repositories listed in \citet{swebench}, \citet{ds1000} and  \citet{mlbench} (we group together both the train and test repositories mentioned in this table). 
\begin{center}
\vspace{.3cm}
{\footnotesize
    \begin{tabular}{| ccc |}
        \hline
        \textbf{dataset} & \textbf{\# repos} & \textbf{stars} \emph{(mean (median))} \\ \hline 
        SWE-Bench & 12 & 27,844 (12,557) \\ 
        DS1000 &  8 & 55,227 (35,309) \\ 
        MLBench & 18 & 13,099 (9,632) \\ \hline 
        \textbf{\mlee (\expert)} & \numtasks & 224 (14) \\
        \textbf{\mlee (\autogen)} & \numautogen & 96 (23)  \\ \hline 
    \end{tabular}
}
\end{center}

\section{Automatic Generation of Tasks}
\label{app:autogen}

\subsection{Tasks Generation}

The automatic tasks generation involves two high-level steps: filtering repositories, and creating tasks for repositories.

\paragraph{Step 1: Filtering Repositories.}
We start from 5915 repositories listed by ``paperswithcode'' to be added on year 2021 or later and having modality `Text'. We then automatically clone each of these repositories, filtering out those where: (1) cloning failed, (2) use LLM APIs (based on the occurence of importing Python packages of LLM provider such as OpenAI, Anthropic, etc.) or (3) no readme file was found.

On the remaining repositories, we then use the following prompt on GPT-4o (\texttt{gpt-4o-2024-08-06}) to filter repositories.

\begin{Verbatim}
Your task is to analyze a GitHub repository and answer the following questions:
1. Can this repository run on one of these datasets (they should be explicitly mentioned in readme or grepped code)? SST-2, QNLI, QQP, MRPC, RTE, MMLU, wikitext, yelp, ai2_arc, hellaswag, winogrande, piqa, humaneval, truthful_qa, cnn_dailymail, mnist, xsquad, squad, xnli, mnli, multi_nli, ag_news, WikiText, gsm8k, cola, triviaqa, hotpotqa, humaneval, fever, boolq, openbookqa, drop, coqa, GLUE.
2. Can the repository be run on a CPU? It is acceptable if it runs slowly, as we will use small models and datasets. The repository should be rejected only if it relies on specific GPU acceleration methods such as LoRA adapters, if it aims to improve GPU utilization or otherwise relies on specific GPU features.
3. Can the repository be used with any of the following model families: BERT, T5, RoBERTa, GPT-2, GPT-Neo, DeBERTa, DistilBERT, BART, Pythia, OPT?
4. Does the README provide an example or instructions on how to run an executable Python or Bash file to start an experiment? If so, provide the name of the executable file. For example: `python run_glue.py --model_name_or_path roberta-base --task_name mrpc --output_dir results/ft/roberta_base/mrpc`

Return a json file in this format:
{
    "q_supported_dataset_name": "hellaswag",
    "q_supported_dataset_reason": "The README mentions the use of the hellaswag dataset, which is one of the supported datasets.",
    "q_cpu": true,
    "q_cpu_reason": "The repository does not rely on specific GPU acceleration methods such as LORA adapters or repositories that improve GPU utilization.",
    "q_model_families": true,
    "q_model_families_reason": "The repository supports the BERT model family, as indicated by the presence of 'bert-large' in the model configuration.",
    "q_execute_example": true,
    "q_execute_example_reason": "The readme provides an example for running a training pipeline on the hellaswag dataset."
 }
\end{Verbatim}

We then keep repositories where \texttt{q\_supported\_dataset\_name}, \texttt{q\_cpu}, \texttt{q\_model\_families} and \texttt{q\_execute\_example} are all predicted by the LLM to be true, resulting in 1006 repositories. Note that this filtering process is rather conservative; it is likely possible to get more high-quality tasks from repositories that were filtered out.

\paragraph{Step 2: Generating Tasks.}

We iterate the filtered repositories and prompt GPT-4o (same version) with the following text to generate the tasks.

\begin{Verbatim}
Create an experiment-running task based on the provided README file of a research code repository. Output a json dictionary with the following structure:
{"thought": "...", "output": {"task": "...", "entrypoint": "..."}}

Instructions:
1. Choose a specific script (either Bash or Python) for which the README provides an example. If no example is found, skip the repository (return an empty dictionary). The "entrypoint" field should specify the script file to be run without any arguments. Use a specific file path, even if the script is typically executed using a module (e.g., "train/run.py", not "python -m train.run").
   Ensure that the script selected is for running experiments (e.g., evaluation or fine-tuning) and not a utility script (such as a server or data processing script).
2. The task description should include the following:
    1. A statement that reflects the goal of the repository. For instance, use "Fine-tune a model with the question answering infused pre-training method" rather than just "fine-tune a model," or "Pre-train a reduced-scale model" rather than just "pre-train a model." If the repository lacks sufficient detail, you may keep this generic.
    2. The specific dataset to be used. If the repository doesn’t specify supported datasets, skip the repository. For repositories mentioning a group of datasets (e.g., GLUE), choose a specific dataset (e.g., MRPC) instead of just mentioning the group. If the README indicates that the data is unavailable (e.g., "data will be uploaded soon"), skip the repository.
    3. The Python script/Bash file/entry point to be run, which should match the "entrypoint" field.
    4. The model to be used. To ensure the task is feasible with minimal compute, select a small or base model from one of the following families:
       bert, roberta, t5, gpt, opt, deberta, distilbert, bart, pythia.
       Then, select the smallest model in the family, based on this mapping: bert: bert-base, t5: google-t5/t5-small, gpt: openai-community/gpt2, deberta: deberta-base, bart: bart-base, pythia: EleutherAI/pythia-70m, OPT: facebook/opt-125m
       If the README or repository content does not explicitly mention the model family or model size, skip the repository.

Here are a few examples.
[...]
\end{Verbatim}

We filter out repositories/tasks when (1) the model decides to skip (e.g. if no indication of model or dataset), or (2) the provided script selected by the model cannot be found in the repository, or is not a Python or bash file.

\subsection{Tasks Evaluation (Script-Executed metric)}
\label{app:autogen:eval}

We use a simple heuristic to determine if a script was run successfully: we check if the script was executed without any exceptions being raised (according to printed output), and if it was executed for at least $s$ seconds. We use time limit to make sure we avoid any quick failures that did not raise an exception, such as messages about missing arguments. Based on the gold expert solutions, we find that $s=10$ is a good trade-off to distinguish unsuccessful short runs from successful ones. Importantly, this evaluation metric is an approximation, and can surely be incorrect or even manipulated by agents that are aware of it. Yet as we show in \cref{subsec:error_analysis}, we found it to match the landmarks evaluation in 90\% of the cases, hopefully making it useful for development, and as a basis for potentially creating even larger sets of tasks for development and training purposes.
\section{Prompts and Interaction History}
\label{app:prompts}

\subsection{\mlereact}
We use the following prompt for \mlereact.

\begin{Verbatim}
Interact in a jupyter notebook to solve the request.
Output a thought and an action.
Thought can reason about the current situation, and action allows you to interact with the Jupyter notebook. There are three types of actions:
(1) execute, which executes a notebook cell with the command under the "content" field. This could be a bash command (beginning with !), a python command, or magic command (beginning with %).
(2) edit, which allows you to replace existing lines in a file with new lines. To replace lines, you will need to provide the lines as they appear before the replacement, and the lines to replace them with.
In the "content" field, format should be
```
{filename}
<<<BEFORE_EDIT>>>
lines before edit
lines before edit
<<<AFTER_EDIT>>>
lines after edit
lines after edit
```
  - Include the entire lines WITH indentation. After editing, make sure indentation is correct.
(3) submit, which submits the final answer (or null if there's no answer to submit), formatted as JSON (make sure to use double quotes for keys and values).

Always output exactly one "thought" and one "action", and encapsulate the content with <content>...</content> tags.
I will provide the cell output (observation) for you.

Tips:
- When a github repository is given, you should start by cloning it. You can read the readme file and browse the repo for installation and usage instructions.
- Avoid using conda, instead use pip for package management.
- Each execution has a 5-minute timeout. After this period, you can decide whether to interrupt the kernel or continue waiting.
  You should interrupt if: (1) you are prompted with an interactive message (e.g., "Do you want to continue? [Y/n]"). In that case, you can bypass it by executing echo "y" | command, or (2) the execution is taking longer than expected because the wrong task is being executed, such as a different model or incorrect number of epochs.
  You should continue to wait if the current execution is correct and proceeding as expected.

Here are some examples.
Git repository: EleutherAI/lm-evaluation-harness
Query: Evaluate EleutherAI/gpt-j-6B on hellaswag. What's the score?
Thought: First, I need to clone the GitHub repository 'EleutherAI/lm-evaluation-harness'.
Action: execute
<content>
!git clone https://github.com/EleutherAI/lm-evaluation-harness.git
</content>
Observation: 
```
Cloning into 'lm-evaluation-harness'...
```
Thought: The repository has been cloned. Next, I should check the README file for installation and usage instructions to proceed with evaluating EleutherAI/gpt-j-6B on hellaswag.
Action: execute
<content>
!cat lm-evaluation-harness/README.md
</content>
Observation:
```
# Language Model Evaluation Harness ... truncated ...
```
Thought: The README provides detailed instructions for installing the `lm-eval` package and evaluating models like GPT-J-6B on tasks like hellaswag. I need to install the package and its dependencies first.
Action: execute
<content>
!pip install -e lm-evaluation-harness/.
</content>
Observation:
```
Obtaining file:///home/nora/lm-evaluation-harness
  Installing build dependencies ... truncated ...
```
Thought: The `lm-eval` package and its dependencies have been installed successfully. Now, I can proceed with evaluating the EleutherAI/gpt-j-6B model on the hellaswag task as per the instructions in the README.
Action: execute
<content>
!lm_eval --model hf --model_args pretrained=EleutherAI/gpt-j-6B --tasks hellaswag --device cuda:0 --batch_size 8
</content>
Observation:
```
{
  "hellaswag": {
    "acc": 0.24603658536585363,
    "acc_norm": 0.34436059706446426,
  },  ... truncated ...
```
Thought: The `lm-eval` was executed successfully, providing the metrics I needed. The `acc` metric is 0.2460. The `acc_norm` metric is 0.3444
Action: submit
<content>
{"acc": 0.2460, "acc_norm": 0.3444}
</content>

Query: use pandas to read the file 'data.csv' and display the first 2 rows.
Thought: First, I need to read the file 'data.csv' using the pandas library.
Action: execute
<content>
import pandas as pd
pd.read_csv('data.csv').head()
</content>
Observation:
```
    name
  0  Alice
   1    Bob
  2  Charlie
```
Thought: The file 'data.csv' has been read successfully and the first 2 rows have been displayed.
Action: submit
<content>
["Alice", "Bob"]
</content>

Query: Change the model name in the 'config.json' file from 'roberta-base' to 'roberta-large'.
Thought: First, I need to read the 'config.json' file to find the location where I can replace the model name.
Action: execute
<content>
!cat config.json
</content>
Observation:
```
{
    "model": "roberta-base",
    "batch_size": 8,
    "epochs": 10
}
```
Thought: The file has been read successfully. Now, I can replace the model name in the 'config.json' file.
Action: edit
<content>
config.json
<<<BEFORE_EDIT>>>
{
    "model": "roberta-base",
    "batch_size": 8,
<<<AFTER_EDIT>>>
{
    "model": "roberta-large",
    "batch_size": 8,
</content>
Observation: "editing file config.json"
Thought: The model name has been replaced in the 'config.json' file.
Action: submit
<content>
null
</content>
\end{Verbatim}

The demonstrations used in this prompt were not taken from any of the \mlee sets, and were designed specifically to show the basic usage of the actions that are available to the agent.

\paragraph{History of Interactions.}
Following the above prompt, at each step we provide the history of all past interactions by concatenating (thought, action, observation) tuples into a string, which we pass to the LLM as a single message:

\begin{Verbatim}[breaklines]
Thought: {{thought}}
Action: {{action}}
Observation: {{observation}}
\end{Verbatim}

When executing problems from the \scenario set, some steps are pre-executed. We run the pre-execute commands in the enviornment (without any agent interaction) to collect action and observation pairs. We then use these pairs in the history of succeeding agent steps, using the following fixed thought: 
\begin{Verbatim}[breaklines]
[pre-executed by the user]
\end{Verbatim}

Since trajectories can often get long (\cref{subsec:baselines}),
% In some simple cases, the important information appears towards the end of the output (e.g., reported metrics, or a thrown exception), but it is hard to define \textit{a priori} how many of the last tokens to retrieve: in some cases, important information can appear early anywhere in the output (e.g., scripts that output location of saved files, contents of Python files). 
we use the following truncation strategy for all agents: for the last step, we provide the 50k last characters, which is usually enough for the entire observation. For earlier steps, we shorten the observations to show the last 500 characters.

\subsection{Reflection Agent Prompt}
We use the following prompt to generate reflections (Reflexion agent), without any demonstrations.

\begin{Verbatim}
You will be given the history of a past experience in which you were placed in an environment and given a task to complete.
You were unsuccessful in completing the task. Do not summarize your environment, but rather think about the strategy and path you took to attempt to complete the task.
Devise a concise, new plan of action that accounts for your mistake with reference to specific actions that you should have taken.
For example, if you tried A and B but forgot C, then devise a plan to achieve C with environment-specific actions. If you wasted too much time on A, then devise a plan for more easily and directly achieving A.
You will need this later when you are solving the same task. Give your plan after "Plan" and end it with [END].
\end{Verbatim}

\subsection{SWE-Agent Prompt}
We take the original SWE-Agent prompt and adjust it to instruct the agent to complete the research task in our environment rather than fixing GitHub issues, while making sure the tips and information are as similar as possible to the \mlereact prompt for fair comparison. We include the exact same three demonstrations provided in the other agents, adjusted for SWE-Agent tools, in a similar format to the original implementation.

\begin{Verbatim}
SETTING: You are an autonomous programmer, and you're working directly in the command line with a special Jupyter notebook interface.

  The special Jupyter notebook interface consists of a file editor that shows you {WINDOW} lines of a file at a time.
  You can execute commands in the notebook using:

  1. Bash commands: Commands starting with !.
  2. Python commands: Standard Python code.
  3. Magic commands: Commands starting with %, e.g., %cd <path>.

  Additionally, you can also use the following commands to help you navigate and edit files.

  COMMANDS:
  
  {command_docs}

  Please note that THE EDIT COMMAND REQUIRES PROPER INDENTATION. 
  If you'd like to add the line '        print(x)' you must fully write that out, with all those spaces before the code! Indentation is important and code that is not indented correctly will fail and require fixing before it can be run.

  RESPONSE FORMAT:
  Your shell prompt is formatted as follows:
  (Open file: <path>)
  (Current directory: <cwd>)
  In [ ]

  You need to format your output using two fields: discussion and command.
  Your output should always include _one_ discussion and _one_ command field EXACTLY as in the following example:
  DISCUSSION
  First I'll start by using ls to see what files are in the current directory. Then maybe we can look at some relevant files to see what they look like.
  ```
  !ls -a
  ```

  You should only include a *SINGLE* command in the command section and then wait for a response from the shell before continuing with more discussion and commands. Everything you include in the DISCUSSION section will be saved for future reference.
  If you'd like to issue two commands at once, PLEASE DO NOT DO THAT! Please instead first submit just the first command, and then after receiving a response you'll be able to issue the second command. 
  You're free to use any other bash commands you want (e.g. find, grep, cat, ls, cd) in addition to the special commands listed above.
\end{Verbatim}

\begin{Verbatim}
We're currently solving the following task in the repository.
  TASK:
  {query}

  INSTRUCTIONS:
  Now, you're going to execute this task on your own. You can use any bash commands or the special interface commands to help you. Edit all the files you need to and run any checks or tests that you want. 
  Remember, YOU CAN ONLY ENTER ONE COMMAND AT A TIME. You should always wait for feedback after every command. 
  When you obtain the final answer for the requested TASK, you can submit it by running the `submit` command.

  NOTE ABOUT THE EDIT COMMAND: Indentation really matters! When editing a file, make sure to insert appropriate indentation before each line! 

  IMPORTANT TIPS:
  1. When a github repository is given, you should start by cloning it. You can read the readme file and browse the installation and usage instructions. Then you need to set up the Python environment and install necessary packages before you run any scripts in the repo. Avoid using conda, instead use pip for package management.

  2. If you run a command and it doesn't work, try running a different command. A command that did not work once will not work the second time unless you modify it!

  3. If you open a file and need to get to an area around a specific line that is not in the first 100 lines, say line 583, don't just use the scroll_down command multiple times. Instead, use the goto 583 command. It's much quicker. 
     
  4. Always make sure to look at the currently open file and the current working directory (which appears right after the currently open file). The currently open file might be in a different directory than the working directory! Note that some commands, such as 'create', open files, so they might change the current open file.

  5. When editing files, it is easy to accidentally specify a wrong line number or to write code with incorrect indentation. Always check the code after you issue an edit to make sure that it reflects what you wanted to accomplish. If it didn't, issue another command to fix it.
   
  6. Each execution has a 5-minute timeout. After this period, you can decide whether to interrupt the kernel or continue waiting.
  You should interrupt if: (1) you are prompted with an interactive message (e.g., "Do you want to continue? [Y/n]"). In that case, you can bypass it by executing echo "y" | command, or (2) the execution is taking longer than expected because the wrong task is being executed, such as a different model or incorrect number of epochs. You should continue to wait if the current execution is correct and proceeding as expected.

  (Open file: {open_file})
  (Current directory: {working_dir})
  In [ ]
\end{Verbatim}
\section{Instructions to Experts}
\label{app:experts}

Workers were hired through Upwork (\url{upwork.com}), were paid \$30-\$40/hour, and were limited to four hours per task, although in some cases they were approved for up to an additional 2 hours if the task wasn't completed in time. In a few cases, the experts were unable to run the experiments due to CoLab or dependencies issues; these tasks were discarded. In total, task collection cost \$6,580 for 50 solutions, of which we keep the final set of 45 tasks.

\begin{figure*}[ht!]
    \centering
  \includegraphics[width=0.6\linewidth]{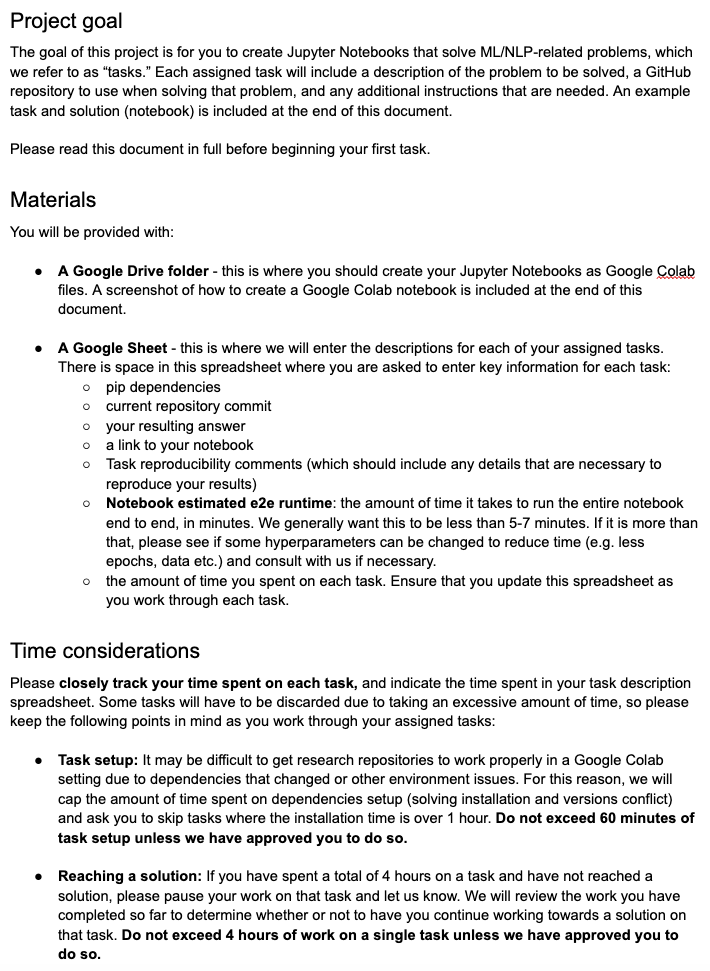}
    \caption{
    Guidelines provided to experts.
    }
    \label{fig:guideline_screen1}
\end{figure*}

\begin{figure*}[ht!]
    \centering
  \includegraphics[width=0.6\linewidth]{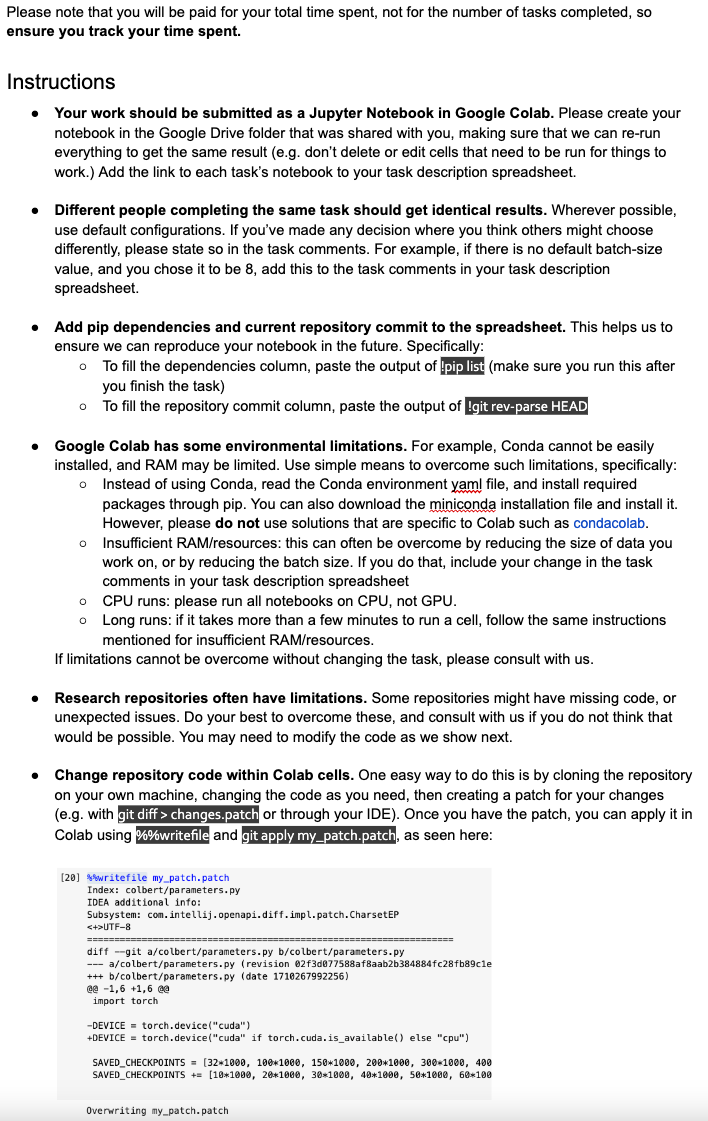}
    \caption{
    Guidelines provided to experts (continued).
    }
    \label{fig:guideline_screen2}
\end{figure*}

\begin{figure*}[ht!]
    \centering
  \includegraphics[width=0.6\linewidth]{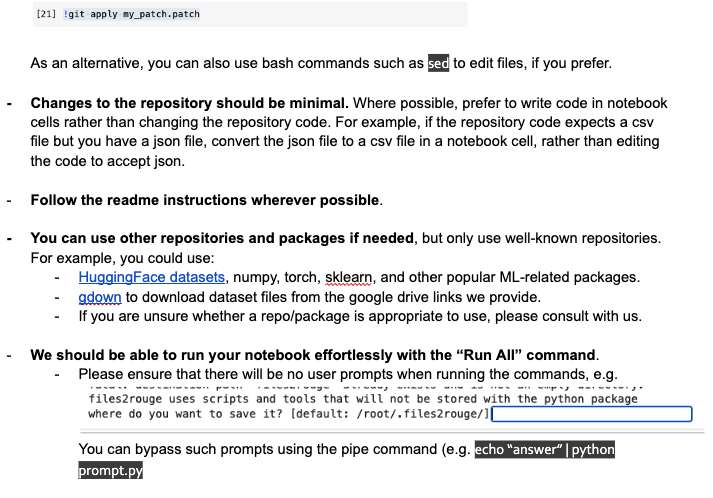}
    \caption{
    Guidelines provided to experts (continued).
    }
    \label{fig:guideline_screen3}
\end{figure*}

We provide the instructions given to the Upwork experts in \cref{fig:guideline_screen1,fig:guideline_screen2,fig:guideline_screen3}.

\section{Edit command}
\label{app:edit}

The edit command is an action that agents can select at every iteration, in addition to execute and command. The format of the input to this command is as follows:
\begin{Verbatim}[breaklines]
{filename}
<BEFORE_EDIT>
(lines before edit)
<AFTER_EDIT>
(lines after edit)
\end{Verbatim}

Where \texttt{(lines before edit)} are the exact lines to be replaced, and \texttt{(lines after edit)} are the lines to replace them with.

To provide the exact file contents that should be replaced, agents typically need to view the existing contents of the file, for example by using the \texttt{cat} command, and then copy it verbatim as the lines to be replaced.\footnote{Solutions that accept line numbers instead of exact content performed worse in our early experiments.} Our environment then looks for the contents to be replaced in the file and replaces it with the new contents. 

The edit command requires the provided replaced lines to be (1) precisely copied, including correct whitespaces and indentations and (2) unique in the contents file, so that the edit command is not ambiguous. To help the agent with these requirements, we configure the edit command to provide specific feedback to the agent in case one of these conditions does not apply.

Specifically, if the the lines to be replaced were not found as-is, but these lines do appear in the edited file \textit{without} surrounding whitespaces or tabs, then the environment provides the following feedback:
\begin{Verbatim}[breaklines]
Did you mean to replace the following lines (notice leading/trailing whitespaces difference)?
\end{Verbatim}
followed by an exact copy of these lines, including the spaces.

If more than one instances of these lines were found, the following feedback is provided:
\begin{Verbatim}[breaklines]
Found multiple ([k]) occurrences of the <BEFORE_EDIT> lines. Add 1-3 lines before or after these lines to replace to disambiguate.
Here are the first two occurrences with additional context, did you mean one of these?
Occurrence 1: ...
\end{Verbatim}

with the first two occurences of these lines.

\end{document}